\documentclass{article}

\usepackage[final]{neurips_2023}

\usepackage[utf8]{inputenc} 
\usepackage[T1]{fontenc}    
\usepackage{hyperref}       
\usepackage{url}            
\usepackage{booktabs}       
\usepackage{amsfonts}       
\usepackage{nicefrac}       
\usepackage{microtype}      
\usepackage{xcolor}         

\usepackage[utf8]{inputenc} 
\usepackage[T1]{fontenc}    
\usepackage{url}            
\usepackage{booktabs}       
\usepackage{amsmath}
\usepackage{amsfonts}       
\usepackage{nicefrac}       
\usepackage{microtype}      
\usepackage{color,xcolor,colortbl}
\usepackage{accents}
\usepackage{pgffor}
\usepackage{subcaption}
\usepackage{enumerate}
\usepackage{multirow}
\usepackage{wrapfig}
\usepackage{ifthen}
\usepackage{caption}
\usepackage{float}
\usepackage{graphicx}
\title{Single Image Reflection Removal with Patch Reflectance Prior}
\author{
Dongshen Han\textsuperscript{1} \qquad Heechan Yoon\textsuperscript{1} 
Hyukmin Kwon\textsuperscript{2} \qquad Hyun-Cheol Kim\textsuperscript{2}  \\ \qquad Hyon-Gon Choo\textsuperscript{2}
\qquad Seungkyu Lee\textsuperscript{1} \qquad  Chaoning Zhang\textsuperscript{1} 
\\
$^1$Kyunghee University\\
$^2$Electronics and Telecommunications Research Institute\\
{\tt\small \{han-0129, seungkyu\}@khu.ac.kr}}

\begin{document}

\maketitle

\begin{abstract}
Single Image Reflection Removal (SIRR) in real-world images is a challenging task due to diverse image degradations occurring on the glass surface during light transmission and reflection. Many existing methods rely on specific prior assumptions to resolve the problem. In this paper, we propose a general reflection intensity prior that captures the intensity of the reflection phenomenon and demonstrate its effectiveness. To learn the reflection intensity prior, we introduce the Reflection Prior Extraction Network (RPEN). By segmenting images into regional patches, RPEN learns non-uniform reflection prior in an image. We propose Prior-based Reflection Removal Network (PRRN) using a simple transformer U-Net architecture that adapts reflection prior fed from RPEN. Experimental results on real-world benchmarks demonstrate the effectiveness of our approach achieving state-of-the-art accuracy in SIRR.
\end{abstract}

\section{Introduction}
The phenomenon of reflection can be seen everywhere in daily life on window glass surfaces where reflected scene is superimposed on transmitted scene. Frequently, such superimposed reflection of other objects of surroundings works as confusing noise in the capturing quality of transmitted scene image.
To address the problem, reflection removal methods model reflection image as a weighted additive composition of a transmission image (T) and a reflection image (R) as formulated by the following equation \cite{zheng2021single}:
\begin{equation}
\label{eqn:01}
{{\textit{I}}=\textit{g}(\textit{T})+\textit{f}(\textit{R})},
\end{equation}

where $\textit{g}(\cdot)$ and $\textit{f}(\cdot)$ represent diverse image degradations that occur on the glass surface during light transmission and reflection such as image blur, light absorption and refraction, multiple offsets, and depth of field (DoF) \cite{wan2017benchmarking}.  
The presence of unknown content-free variables makes single image reflection removal task ill-posed and highly challenging problem.

To address the problem, different prior assumptions have been utilized for retrieving transmission layer information including image gradient sparsity \cite{levin2002learning} and domain \cite{dong2021location}, ghosting cues \cite{shih2015reflection}, relative smoothness \cite{li2014single}, as well as earlier methods such as handcrafted priors \cite{levin2007user}. Prior assumption reasonably work with strong and dominant transmission image (T) and weak reflection image (R) \cite{li2020single}.
These prior assumptions have been shown to assist the network in recognizing reflection image and roughly locate reflection region to some extent \cite{dong2021location}. However, since the amount of real-world data available for single-image reflection removal (SIRR) is still insufficient, researchers have resorted to creating synthetic SIRR datasets by artificially adding noise and ghosting \cite{zhang2018single, fan2017generic}. This has led to many models being designed and trained heavily relying on the specific priors \cite{zhang2018single, dong2021location, wan2019corrn}. There are also many methods that aim to decompose superimposed image into reflection and transmission images allowing them to supervise each other during the generation process \cite{li2020single, hu2021trash, wen2019single}.

Recently, for the advanced development of SIRR, Lei \emph{et al.} \cite{lei2022categorized} propose a novel and diversity dataset named Categorized, Diverse, and Real-world (CDR). CDR dataset is constructed using diverse glass types under various environmental conditions. It is divide into two parts based on the smoothness of reflection: sharp vs blurry reflections. They emphasize that while many prior-based networks and algorithms generally perform well on specific priors such as blurry reflections, they often exhibit poor performance on the scenes from our daily lives with varying object distances, scales, and natural illumination. 
One of the reasons for the poor performance in SIRR is over-reliance on a specific prior assumption in many prior methods. 
However, in the real world, this assumption is limited in its applicability. For example in in-focus reflection \cite{wan2017benchmarking}, superimposed image appears as a simple addition of two clean images without any additional prior information except for the fact that the resulting image is cluttered. As a result, it causes the network to lose its robustness when it deals with superimposed images observed in real-world scenarios.
Furthermore, as stated in Equation \eqref{eqn:01}, there are many factors in real-world superimposed images that affect their reflection image \cite{wan2017benchmarking}. As a result, separating the reflection and transmission images through mutually supervised training and estimation becomes challenging, which in turn leads to unstable performance in real-world scenarios.

In this paper, to address SIRR problem in real-world scenarios, we propose general type of reflection prior and corresponding SIRR process. We do not specify whether this prior is due to abnormal gradient distribution, blurring, ghosting, or any other effects.
Instead of any predetermined type of reflection prior, we simply define reflection intensity prior as the strength of reflection phenomenon occurring in a local region. 
To this end, we train our network to learn the proportion of reflection and transmission strength.
And then, we design Reflection Prior Extraction Network (RPEN). Our target is to employ a backbone model pre-trained on a large dataset of regular images to recognize the differences between regular images and abnormal superimposed images with reflection.
And then RPEN learns reflection intensity prior at each local region patch separately with SIRR dataset. 
Segmenting superimposed images into small patches and labeling them for prior estimation is very challenging. However, observed reflection intensity is not uniform in a superimposed images and regionally adaptive reflection prior optimizes the performance of SIRR.
Using large size of patch in reflection prior reduces the significance of prior information, but small patch with limited amount of observation is difficult to be trained correctly. 
Finally, based on the reflection intensity prior of each region, we
propose Prior-based Reflection Removal Network (PRRN). For the implementation of PRRN, we employ a simple transformer U-Net\cite{saharia2022image} architecture which is proved to be remarkably successful in various image restoration tasks such as denoising \cite{hoogeboom2022blurring,bansal2022cold,saharia2022image}, deraining \cite{ye2022unsupervised}, despite its simplicity. This architecture is not only simple but also demonstrates excellent performance when utilizing prior information.

Our contributions is summarized as follows:

 $\bullet$ 
We propose reflection intensity prior that is free from any specific prior assumption about the reflection phenomenon. And we demonstrate its effectiveness in Single Image Reflection Removal (SIRR) task. Additionally, we create a reflection intensity prior dataset to train our proposed Reflection Prior Extraction Network (RPEN).

$\bullet$ 
We propose Prior-based Reflection Removal Network (PRRN) that utilizes our reflection intensity prior predicted from RPEN. We validate the effectiveness of our approach on the latest real-world benchmarks. Our experimental results demonstrate that our approach achieves state-of-the-art accuracy.

\section{Related Work}
\label{gen_inst}
\subsection{Prior based Image Processing}
In challenging tasks such as camouflage detection, as well as single-image reflection removal, prior information plays a crucial role.
Traditional methods typically solve it by imposing handcrafted priors \cite{levin2007user} or specific prior information \cite{levin2002learning}, such as the ghosting artifact \cite{shih2015reflection}. 
Levin \emph{et al.} \cite{levin2007user}  proposed a method for removing reflections from images taken through glass surfaces that exploits the characteristics of reflections and backgrounds by applying sparsity priors. However, their approach requires users to provide prior information, such as manually labeling the background and reflection edges, which is a labor-intensive task and may not be reliable in regions with complex textures.
Wan \emph{et al.} \cite{wan2016depth} presented a visual depth-guided method for removing reflections by using Depth of Field (DoF) to label the background and reflection edges. However, this approach has some limitations, such as requiring accurate depth information and potentially being ineffective for images with complex reflections. Shih \emph{et al.} \cite{shih2015reflection} utilized the common prior of ghosting artifacts in reflections for their reflection removal task. They modeled the ghosted reflection using a double-impulse convolution kernel and automatically estimated the spatial separation and relative attenuation of the ghosted reflection components. Arvanitopoulos \emph{et al.} \cite{arvanitopoulos2017single} proposed an approach to remove reflection artifacts from images captured through glass windows by incorporating gradient prior assumptions based on physical properties into optimization schemes. However, these prior assumptions are too specific to handle complex real-world reflection scenarios. On the other hand, manually labeling the regions with prior information is a labor-intensive task and may not be reliable in regions with complex textures.

Deep learning has demonstrated impressive performance in various computer vision tasks, and its comprehensive modeling capability can also be applied to obtain important prior information for challenging image processing tasks, including reflection removal, segmentation of glass and mirrors, and other types of image-based camouflage detection\cite{zhang2021depth}. Wan \emph{et al.} \cite{wan2018region} propose a region-aware reflection removal method that uses deep learning to automatically detect and process regions with and without reflections. The method utilizes content and gradient priors for joint restoration of missing information.
Dong \emph{et al.} \cite{dong2021location} use LSTM for refinement and employed Laplacian kernels of different sizes to obtain prior information about the reflection region. Based on this information, it generated a confidence map to guide the neural network in segmenting the projected image and the reflection image. Lin \emph{et al.} \cite{lin2022exploiting} and Guan \emph{et al.} \cite{guan2022learning} demonstrate that by exploiting the semantic prior information of objects in the environment where glass or mirror objects were located, they were able to guide neural networks to more accurately locate these challenging camouflaged objects in images.

\subsection{Single Image Reflection Removal}
SIRR operates on an image $\textit{I}$ that is typically a combination of a transmission layer $\textit{T}{'}$ and a reflection layer $\textit{R}{'}$, and many methods \cite{hu2021trash,wan2019corrn,li2020single,dong2021location} supervise the generation of $\textit{T}{'}$ by monitoring $\textit{R}{'}$.
Hu \emph{et al.} \cite{hu2021trash} propose a dual-stream decomposition network for reflection and transmission separation, which enhances communication between the two branches by enforcing block-wise interaction. This approach, called Your Trash is My Treasure (YTMT), enables the transmission branch to utilize information from the reflection branch, allowing for the removal of useless information in the reflection branch and improving the quality of transmission generation. 
Wan \emph{et al.} \cite{wan2019corrn} propose a statistic loss to remove strong reflections in local regions by considering gradient level statistics, building upon their use of the reflection branch to supervise the training of the transmission branch. The reflection branch in dual-branch networks trained on real-world superimposed images is affected by the ill-posed nature of the problem, especially in the case of strong reflections where the degradation of the reflection layer from the superimposed image is more complex. This can lead to the transmission branch being corrupted by noise transmitted from the reflection branch during training, resulting in failed transmission layer separation.
Li\emph{et al.} \cite{li2020single} propose the IBCLN method, which is a recurrent LSTM\cite{hochreiter1997long}-based network that utilizes temporal properties to progressively separate and refine the predicted reflection and transmission layers. However, this iterative refinement approach is time-consuming.

\begin{figure*}[t!]
\centering
\begin{subfigure}{\linewidth}
\centering
\includegraphics[scale=0.45]{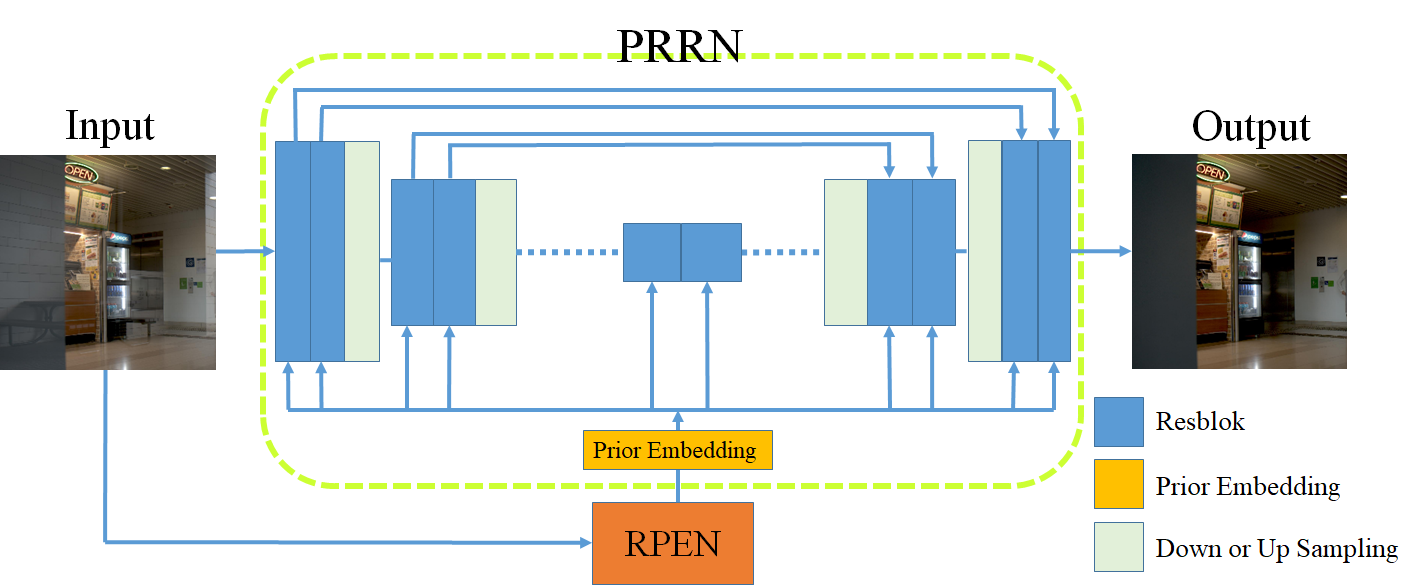}
\end{subfigure}
\caption{The pipeline of our proposed process.}
\label{fig:all}
\end{figure*}

\section{Method}
In this section, we describe proposed single image reflection removal (SIRR) process for diverse real-world reflection images. First, single input image is divided into patches to compute the reflection intensity prior for each patch. Secondly, we train our Reflection Prior Extraction Network (RPEN) to estimate the reflection intensity prior for each region in input superimposed image. Finally, the prior knowledge is fed to our Prior-based Reflection Removal Network (PRRN) to region-adaptively remove reflection noise restoring transmission image as shown in Figure \ref{fig:all}.

\begin{figure*}[t!]
\centering
\begin{subfigure}{\linewidth}
\centering
\includegraphics[scale=0.52]{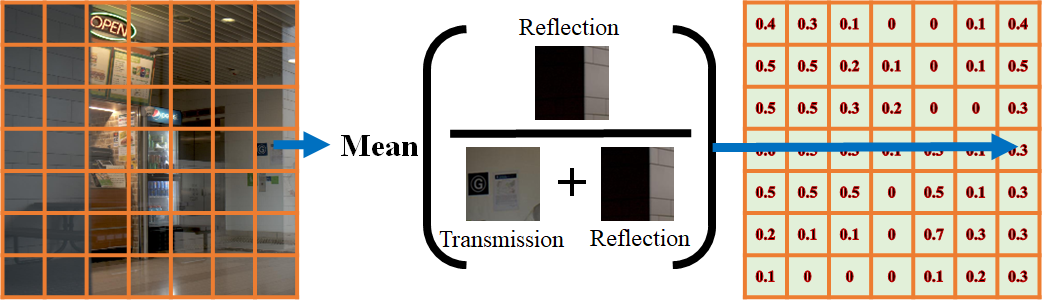}
\end{subfigure}
\caption{Patch segmentation and reflection intensity prior calculation for each patch.}
\label{fig:pathsize}
\end{figure*}
\subsection{Reflection Intensity Prior}
The definition of reflection intensity in an image always has been somewhat ambiguous. In the CDR benchmark \cite{lei2022categorized}, reflection intensity of a superimposed image is briefly introduced during the image categorization process and formulated as follows.
\begin{equation}
\label{eqn:reflect_intensity}
{Reflection Intensity=\frac{\textit{Mean}(R)}{\textit{Mean}(T)}},
\end{equation}
where reflection intensity is defined as the ratio of the reflection layer to the transmission layer of an image with reflection. This is consistent with the way human observe reflection phenomena, where image pollution caused by reflection is not only related to the reflection layer but also closely related to the transmission layer. When the transmission layer is weak, even weak reflection makes it difficult for humans to recognize the transmission layer from an image with reflection (image with superimposed reflection and transmission).
However, this calculation often results in infinite value in the cases of strong reflections, which is not suitable for practical use and computation. To improve the accuracy and avoid the limitation of infinite values of the previous definition, we redefine the reflection intensity as the ratio between the mean of the reflection layer and the sum of the mean values of the reflection and transmission layers, as formulated in the following equation. 
\begin{equation}
\label{eqn:reflect_intensity1}
{Reflection Intensity=\frac{\textit{Mean}(R)}{\textit{Mean}(R)+\textit{Mean}(T)}},
\end{equation}
New definition of reflection intensity allows us to obtain normalized score between the range of 0 to 1. When the reflection intensity approaches 1, it indicates that an image has higher proportion of reflection with severe reflected contamination.

We have observed that reflections occur non-deterministically in certain regions of superimposed image \cite{zhang2018single}. Therefore, we extract the reflection intensity priors on a per-segment patch basis as shown in Figure \ref{fig:pathsize}. But many characteristics of reflections require a large receptive field to capture and define them \cite{mei2020don}. We demonstrate in Figure \ref{fig:patcherror} that using smaller patch leads to larger unit errors. Furthermore, as shown in Table \ref{table:varypatchssim}, we observe that PRRN achieves better SIRR performance as the patch size becomes smaller when using the truth prior as input. 
To address this, we calculate the reflection intensity of entire image and further refine it by dividing it into 49 (7x7) patches.


\subsection{Prior-based Reflection Removal Network}
\vspace{-2mm}
We use a simple transformer U-Net architecture and prior embedding module to build PRRN. Despite its simplicity, U-Net architecture has been proven to be highly successful in various image processing tasks such as denoising and deraining. By leveraging prior information such as the Gaussian noise level, PRRN not only maintains a simple architecture but also exhibits outstanding performance.

The prior embedding module conducts prior encoding to convert reflection intensity prior of each patch into reflection intensity features. This is achieved by referring to the transformer as follows.
\begin{equation}
\label{eqn:downsample}
\begin{aligned}
PriorEncoding_{(prior, 2i)} &= \sin(prior / 10000^{2i /64}) \\
PriorEncoding_{(prior, 2i+1)} &= \cos(prior / 10000^{2i /64}).
\end{aligned}
\end{equation}
$prior$ refers to reflection intensity prior of each patch, and $i$ is the dimension.
$PriorEncoding(\cdot)$ encodes reflection intensity prior $\textit{I}^{\textit{P}\times\textit{P}}$ into reflection intensity features $\textit{I}^{\textit{P}\times\textit{P}\times\textit{64}}$.

UNet structure is a simple encoder-decoder structure with downsampling and upsampling. A transformer is added to the end of the encoder part. To enhance prior  performance, we add two Resblocks prior to upsampling and downsampling operations. These Resblocks are influenced by our reflection intensity prior information which guides the network to remove reflection region-adaptively. 

As a module utilizing ${\textit{I}^{\textit{P}\times\textit{P}\times\textit{64}}  }$, the Resblock mainly consists of two convolutional blocks, FWA (Feature Wise Affine) and a residual convolution.
Input feature ${\textit{F}_{\textit{in}}}$ is processed by a convolutional block to generate ${\textit{F}'{\textit{in}}}$, which is then passed to both FWA and the residual convolution. FWA takes in the input feature  ${\textit{F}'_{\textit{in}}}$ and  ${\textit{I}^{\textit{P}\times\textit{P}\times\textit{64}}  }$. 
${\textit{I}^{\textit{P}\times\textit{P}\times\textit{64}}  }$ is transformed into a specified dimension of ${\textit{I}^{\textit{P}\times\textit{P}\times\textit{C}}  }$ using a fully connected layer followed by nearest-neighbor interpolation in the spatial domain to convert reflection intensity features into the same size as the input feature ${\textit{F}'_{\textit{in}}}$, ${\textit{I}^{\textit{H}\times\textit{W}\times\textit{C}}  }$. 
After adding ${\textit{F}'_{\textit{in}}}$ and ${\textit{I}^{\textit{H}\times\textit{W}\times\textit{C}}}$, we pass them through the final convolutional block and combine the result with the residual convolution output to obtain the final output.

\subsection{PRRN with true Prior} 
\vspace{-2mm}
To validate the effectiveness of our reflection intensity prior, we first partition the image into different patch resolutions (1$\times$1, 7$\times$7, 14$\times$14, 28$\times$28) as shown in Figure \ref{fig:pathsize} to obtain reflection intensity prior maps at different patch resolutions. Then, we use prior embedding to convert them into prior features and train PRRN with each of them. As shown in Table \ref{table:varypatchssim}, we find that as the patch resolution increase (then each patch size decrease) and the reflection intensity prior of input image is subdivided more, reflection removal performance of PRRN on CDR benchmark is improved. 
Figure \ref{fig:patchsize} demonstrates that PRRN has the capacity to remove sharp reflection phenomena such as in-focus reflection \cite{wan2017benchmarking} as the fine-grained reflection intensity prior truth is fed. This shows that obtaining sub-regional reflection intensity prior is crucial for improved PRRN performance.

\subsection{Reflection Prior Extraction Network}
\vspace{-2mm}
To obtain the reflection intensity prior, we construct an RPEN (Reflection Prior Extraction Network). We employ ResNext101 pre-trained on ImageNet \cite{deng2009imagenet} as backbone of RPEN to provide the understanding of regular images without reflections. And then, RPEN is fine-tuned with reflection images to learn regional reflection intensity from superimposed images. We add a fully connected layer on top of the last output layer of the backbone to calculate reflection intensity. Then, we employ a cascaded approach to further refine and separate the reflection intensity of each patch by using ASPP (Atrous Spatial Pyramid Pooling) \cite{chen2018reverse} on high-level features. ASPP considers global relationships between the reflection intensity priors of each patch.
Our RPEN is trained independently in an end-to-end manner prior to the training of PRRN so that RPEN serves as local regional guide for adaptive reflection removal in PRRN.

\section{Experiments}
\subsection{Implementation Details}
\vspace{-2mm}
In our experimental evaluation, RPEN is trained first to handle the extraction of reflection intensity prior and then PRRN is trained with the guidance of fixed RPEN. 
To train RPEN, a training dataset consisting of ground truth pairs of reflection and transmission images is necessary to compute the reflection intensity prior. However, PRRN does not rely on ground truth reflection images as it can directly perform reflection removal tasks using the prior knowledge acquired from RPEN.
We train RPEN on the combination of 7,643 synthetic images \cite{zhang2018single} and 614 real images from CDR training set. The synthesized images are created by selecting images from PASCAL VOC dataset \cite{everingham2010pascal} and superimposing reflection on them using the CEILNet \cite{fan2017generic} to create desired superimposed images.
To address the issue of poor data adaptation caused by inconsistent training methods used in many models for SIRR task and to ensure fair comparison, we train our PRRN using three different training strategies: synthesis \cite{wei2019single}, IBCLN \cite{li2020single} and CDR. Lei \emph{et al.} \cite{lei2022categorized} propose CDR dataset which is recently released and reflects diverse real-world environments. CDR test set is organized with distinct environments such as SRST (sharp reflection and sharp transmission), BRST (blurry reflection and sharp transmission), Non-ghosting, Weak R (weak reflection), Moderate R (moderate reflection), Strong R (strong reflection), and Ghosting. These phenomena enable comprehensive evaluation of model performance in diverse and challenging scenarios. Our primary focus is comparing and conducting ablation experiments on CDR dataset.
Furthermore, we perform additional tests on other datasets to further validate proposed method. It is important to note that our training hyper-parameters are aligned with baseline and the model undergoes extensive training for a substantial number of epochs to ensure its suitability.

Fisrt, for the CDR strategy, we train our model using the CDR trainset and evaluate its performance on the CDR test set. Lei \emph{et al.} \cite{lei2022categorized} use pre-trained models of state-of-the-art methods by default. We not only follow their testing protocol, but also fine-tune or retrain the models that have publicly available training code. This approach allows for a more fair comparison with CDR, as we do not simply rely on their pre-trained models for testing.
Secondly, for the synthesis strategy, we use a dataset that includes 90 pairs of real-world images \cite{zhang2018single} and 7,643 pairs of synthetic images. We compare our trained model with state-of-the-art methods on Real20 \cite{zhang2018single} and three subsets of SIR2 \cite{wan2017benchmarking} benchmarks ensuring fair comparisons.
Finally, we employ the training strategy that IBCLN \cite{li2020single} conducted and evaluate our method on the test data of the Nature dataset introduced by them.
All models are trained on a PC with one 16-core i7-9700K 3.6 GHZ CPU and one NVIDIA GeForce RTX 3090 GPU.

\subsection{Experimental Evaluations}
\vspace{-2mm}
We compare our method with state-of-the-art methods as summarized in Table \ref{table:cdr}, Table \ref{table:synthetic}, and Table \ref{table:ibcln}.
To ensure a fair comparison with CDR benchmark test, we retrain the latest SIRR models, including ERRNet \cite{wei2019single}, IBCLN \cite{li2020single}, YTMT \cite{hu2021trash} on CDR training set.
PSNR and SSIM metrics are utilized for quantitative comparison.
Despite we resize input image to 224x224 for the test of our method, our method consistently outperform other approaches achieving best scores on almost all of the testing datasets. 
In the evaluation on CDR test set in Table \ref{table:cdr}, it is evident that PRRN with RPEN shows promising enhancement in performance over state-of-the-art (SOTA) methods (smallest gain in PSNR is 2.27). 
In the case of SRST set where assumptions like ghosting do not hold and many reflection removal methods experience performance degradation, our method demonstrates exceptional performance outperforming. 
In the case of BRST set where various assumptions hold and most of the previous methods could effectively remove reflections, both PRRN and RPEN shows outstanding performance. This indicates that our method excels in learning and incorporating diverse reflection intensity prior information that is specific to the given phenomena.
Additionally, resizing the test images to 224x224 pixels may result in information loss and hinder the performance in phenomenon with strong reflections.

\begin{figure*}[t!]
\centering\vspace{-4mm}
\begin{subfigure}{\linewidth}
\centering
\includegraphics[scale=0.7]{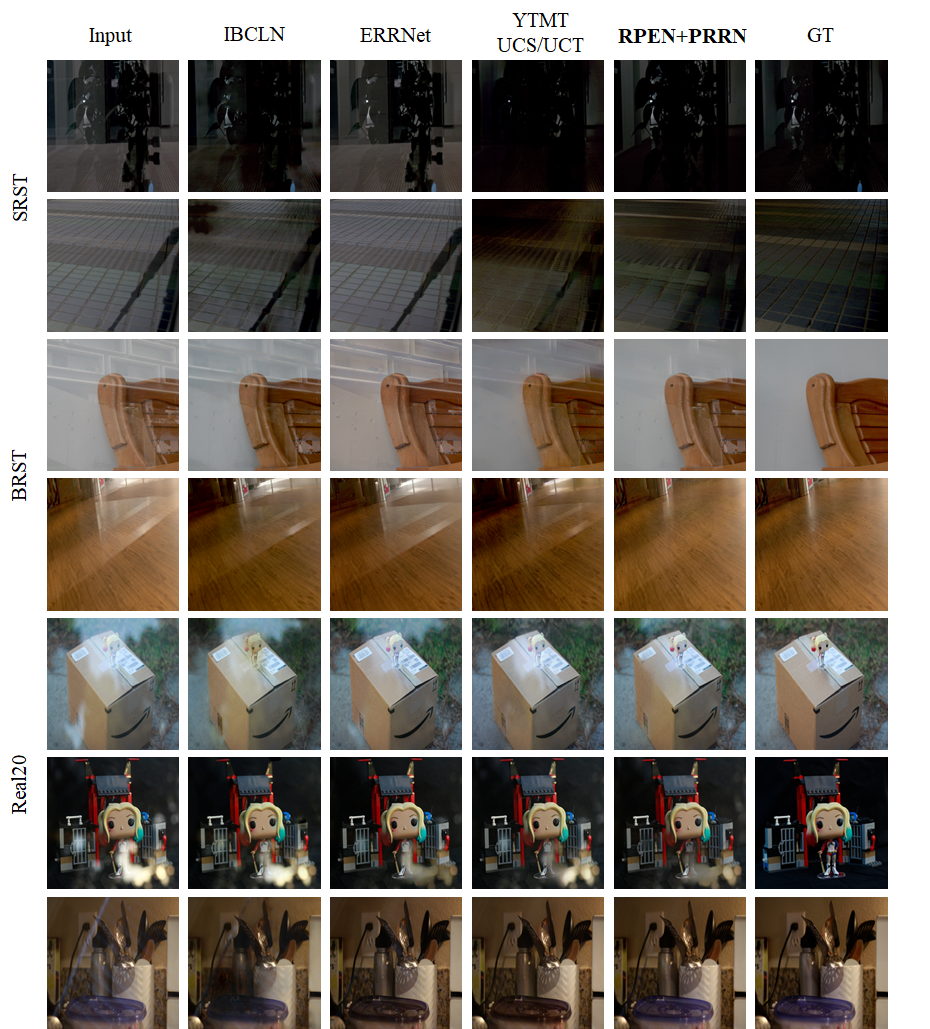}
\end{subfigure}
\caption{Sample qualitative comparison results on CDR and Real20 datasets. Our proposed method demonstrates outstanding performance in sharp reflection environments, such as SRST, and it also shows promising results in strong prior assumption environments, like BRST.}
\vspace{-4mm}
\label{fig:resultimg}
\end{figure*}

In addition to quantitative evaluations, we also conduct visual comparisons in these two environments. As shown in Figure \ref{fig:resultimg}, our method successfully separates reflection image. It is worth to note that YTMT-UCS outperforms YTMT-UCT on CDR dataset and only YTMT-UCS result is shown in the results.
We also test on Synthesis \cite{zhang2018single} and Nature (IBCLN) \cite{li2020single} following the same training strategy. Evaluation results are shown in Table \ref{table:synthetic} and Table \ref{table:ibcln}.
Our method performs reasonably well in removing real-world reflections on the datasets. However, it shows limited performance on postcard benchmark. 
Reflections appeared in postcard images and other real-world reflections mainly on glass surface show different characteristics. Specifically, the degradation function that transforms the reflection image to the superimposed image may vary significantly between these two scenarios. 

\begin{table*}[t]
\centering
\renewcommand{\arraystretch}{1.2}
\begin{tabular}{lcccccccc}
\toprule[1pt]
& \multicolumn{2}{c}{All} & \multicolumn{2}{c}{SRST}& \multicolumn{2}{c}{BRST}& \multicolumn{2}{c}{Non-ghosting}\\
               & {\small PSNR} & {\small SSIM}& {\small PSNR} & {\small SSIM} & {\small PSNR} & {\small SSIM} & {\small PSNR} & {\small SSIM}\\ \hline

\small{CEILNet~\cite{fan2017generic}} & 
\small{17.96} & \small{0.708}  & \small{16.17} & 
\small{0.596} & \small{19.49} & \small{0.802}  & \small{17.24} & \small{0.673}   \\

\small{Zhang et al.~\cite{zhang2018single}} & 
\small{15.20} & \small{0.694}  & \small{13.52} & 
\small{0.590} & \small{16.58} & \small{0.780}  & \small{14.48} & \small{0.662}   \\

\small{BDN~\cite{eccv18refrmv_BDN}} & 
\small{18.97} & {\small{0.758}}& \small{19.04} & 
{\small{0.713}} & \small{19.06} & \small{0.799}  & \small{18.62} & \small{0.733}  \\


\small{Wei et al.~\cite{wei2019single_ERR}} & 
{{\small{21.01}}} & \small{0.762} & {{\small{19.52}}} & 
\small{0.672} & {{\small{22.36}}} & {\small{0.839}} & {\small{20.50}} & \small{0.731}  \\

\small{CoRRN~\cite{wan2019corrn}} & 
{\small{20.22}} & {\small{0.774}}  & {\small{20.32}} & 
{\small{0.699}}  & {\small{20.08}} & {\small{0.838}}  & {\small{20.37}} & \ 
\small{0.750}  \\

\small{IBCLN~\cite{li2020single}} & 
\small{19.85} & {\small{0.764}} & \small{18.33} & 
\small{0.671} & {\small{21.14}} & {\small{0.842}}  & \small{19.23} & {\small{0.735}}   \\

\small{Kim et al.~\cite{Kim_2020_CVPR}} & 
{\small{21.00}} & \small{0.760} & {\small{19.27}} & 
\small{0.676} & {\small{22.61}} & \small{0.833} & 
{\small{20.42}} & \small{0.731} \\
\hline
\small{ERRNet~\cite{wei2019single_ERR}\textsuperscript{*}} &  \small{21.14}
 &  \small{0.762} &  \small{19.51} & 
 \small{0.674} &  \small{22.74} &  \small{0.845} & 
 \small{20.61} &  \small{0.739} \\
\small{IBCLN~\cite{li2020single}\textsuperscript{*}} &  \small{21.12}
 &  \small{0.784} &  \small{19.84} & 
 \small{0.715} &  \small{22.33} &  \small{0.848} & 
 \small{20.44} &  \small{0.760} \\
\small{YTMT-UCS~\cite{hu2021trash}\textsuperscript{*}} &  \small{22.04}
 &  \small{0.774} &  \small{20.89} & 
 \small{0.709} &  \small{22.34} &  \small{0.819} & 
 \small{21.08} &  \small{0.741} \\
 
\hline
\small{PRRN} &  \small\textcolor{blue}{23.44}
 &  \small\textcolor{blue}{0.806} &  \small\textcolor{blue}{22.61} & 
 \small\textcolor{blue}{0.738} &  \small\textcolor{blue}{24.39} &  \small\textcolor{blue}{0.866} & 
 \small\textcolor{blue}{22.92} &  \small\textcolor{blue}{0.784} \\
 
\small{\textbf{PRRN+RPEN}} &  \small\textcolor{red}{24.31}
 &  \small\textcolor{red}{0.815} &  \small\textcolor{red}{23.13}& 
 \small\textcolor{red}{0.748} &  \small\textcolor{red}{25.51} &  \small\textcolor{red}{0.875} & 
 \small\textcolor{red}{23.71} &  \small\textcolor{red}{0.796} \\

  \bottomrule[1pt]
\end{tabular}

\vspace{1mm}

\begin{tabular}{lcccccccc}
    \toprule[1pt]
& \multicolumn{2}{c}{Weak $R$} & \multicolumn{2}{c}{Moderate $R$}& \multicolumn{2}{c}{Strong $R$}& \multicolumn{2}{c}{Ghosting}\\
               & {\small PSNR} & {\small SSIM} & {\small PSNR} & {\small SSIM} & {\small PSNR} & {\small SSIM} & {\small PSNR} & {\small SSIM} \\ \hline

\small{CEILNet~\cite{fan2017generic}} & 
\small{21.34} & \small{0.862}  & \small{17.02} & 
\small{0.685}  & \small{12.06} & \small{0.341}  & \small{20.51} & \small{0.836}   \\

\small{Zhang et al.~\cite{zhang2018single}} & 
\small{17.20} &   \small{0.827}  & \small{15.10}  & 
\small{0.685} & \small{9.33} &\small{0.311}   & \small{17.81} & \small{0.806 }  \\

\small{BDN~\cite{eccv18refrmv_BDN}} & 
\small{21.10} & \small{0.867}  & \small{18.25} & 
\small{0.746} & \small{16.15} & \small{0.485}   & \small{20.20} & \small{0.850}  \\


\small{Wei et al.~\cite{wei2019single_ERR}} & 
{\small{24.89}}& {\small{0.901}} & {\small{19.42}}  & 
\small{0.737} & {\small{17.00}}  & {{\small{0.450}}}  & {{\small{22.80}}} & {\small{0.871}} \\

\small{CoRRN~\cite{wan2019corrn}} & 
\small{20.50} & \small{0.890}  &{\small{21.01}}  & 
{\small{0.768}}  & \small{15.12} & {\small{0.433}}  & \small{19.70} & \small{0.861}   \\

\small{IBCLN~\cite{li2020single}} & 
\small{23.17} & {\small{0.899}} & \small{18.98} & 
{\small{0.752}}  & \small{13.81} & \small{0.395} & {\small{22.07}} & {\small{0.867}}   \\

\small{Kim et al.~\cite{Kim_2020_CVPR}} & 
{\small{25.03}} & \small{0.897} & {{\small{19.66}}} & 
\small{0.740}& {\small{15.25}} & \small{0.431}  & {\small{23.10}} & \small{0.865}  \\
\hline
\small{ERRNet~\cite{wei2019single_ERR}\textsuperscript{*} }&  \small{24.88}
 & \small{ 0.889} &  \small{19.89} & 
 \small{0.750} &  \small{16.09} &  \small{0.464} & 
 \small{23.09} &  \small{0.859} \\

\small{{IBCLN}~\cite{li2020single}\textsuperscript{*}} &  \small{23.76}
 &\small{0.887} &  \small{22.29} & 
 \small{0.773} &  \small{17.37} &  \small{0.514} & 
 \small{23.75} &  \small{0.874} \\
 
 \small{YTMT-UCS~\cite{hu2021trash}\textsuperscript{*}} &  \small{24.34}
 & \small{0.875} &  \small{20.68} & 
 \small{0.746} &  \small\textcolor{red}{18.31} &  \small\textcolor{red}{0.526} & 
 \small{23.74} &  \small{0.855} \\
\hline
\small{PRRN} & \small\textcolor{blue}{26.85}
 &  \small\textcolor{red}{0.906} & \small\textcolor{blue}{22.79} & 
\small\textcolor{blue}{0.809} & \small\textcolor{black}{17.41}  & \small\textcolor{black}{0.493}  & 
\small\textcolor{blue}{25.84} & \small\textcolor{red}{0.892} \\

\small{\textbf{PRRN+RPEN}}& \small\textcolor{red}{28.04}
 &  \small\textcolor{blue}{0.905} & \small\textcolor{red}{23.34} & 
\small\textcolor{red}{0.818} & \small\textcolor{blue}{18.07}  & \small\textcolor{blue}{0.523}  & 
\small\textcolor{red}{26.63} & \small\textcolor{blue}{0.887} \\

  \bottomrule[1pt]
\end{tabular}

\vspace{-1mm}

\caption{
Experimental Comparison on the CDR dataset. The paper provides a detailed analysis. Signal\textsuperscript{*} indicates models that have been retrained by us. The best and second best results are highlighted in \textcolor{red}{red} and \textcolor{blue}{blue}, respectively.}

\label{table:cdr}
\end{table*}

\begin{table}[t]
 \centering
 \vspace{-3mm}
  \scalebox{0.68}
  {
    \begin{tabular}{ccccccccccc}
    \toprule[1pt]
    Datasets  & Metrics   & Zhang \emph{et al.}~\cite{zhang2018single} & BDN~\cite{eccv18refrmv_BDN}& ERRNet~\cite{wei2019single_ERR} & IBCLN~\cite{li2020single} &Lei \emph{et al.}~\cite{lei2020polarized} & YTMT-UCT~\cite{hu2021trash} &\textbf{PRRN+RPEN}  \\ \hline
  \multirow{2}{*}{Real20 (20)}   & PSNR     & 22.55 & 18.41 & {22.89}  & 21.86&22.35  & \textcolor{blue}{23.26}  & \textcolor{red}{23.78}\\
   & SSIM & \small{0.788} & \small{0.726} & \small{0.803}  & \small{0.762} & 0.793&\textcolor{blue}{0.806}&\textcolor{red}{0.838}\\ \hline
  \multirow{2}{*}{Objects (200)}  & PSNR   & 22.68 & 22.72 & {24.87} & 24.87  &23.81&  \textcolor{blue}{24.87}& \textcolor{red}{25.08}\\\
  & SSIM & 0.879  & 0.856 & {0.896}  & 0.893 & 0.882& \textcolor{blue}{0.896}&\textcolor{red}{0.908}\\ \hline 
  \multirow{2}{*}{Postcard (199)} & PSNR  & 16.81  & 20.71 & 22.04  & \textcolor{red}{23.39} & 21.48 & \textcolor{blue}{22.91}&21.01  \\
  & SSIM  & 0.797  & 0.859 & \textcolor{blue}{0.876}  & 0.875 & 0.873 & \textcolor{red}{0.884} &0.856 \\ \hline
  \multirow{2}{*}{Wild (55)} & PSNR & 21.52  & 22.36 & 24.25  & 24.71 & 23.84 & \textcolor{blue}{25.48}&\textcolor{red}{25.48} \\
  & SSIM & 0.832 & 0.830 & 0.853  & 0.886 & 0.866 & \textcolor{red}{0.890} &\textcolor{blue}{0.880}  \\ 
  \bottomrule[1pt]
    \end{tabular}
  }
  \caption{Experimental comparison on four real-world benchmarks using the synthetic strategy. The best results are highlighted in \textcolor{red}{red}, and the second best results are indicated in \textcolor{blue}{blue}.}
  \vspace{-3mm}
    \label{table:synthetic}
\end{table}
\begin{table}[t]
  \centering

  \resizebox{\textwidth}{!}{
      \begin{tabular}{cccccccc}
          \toprule[1pt]
          Metrics & Zhang \emph{et al.}~\cite{zhang2018single} & BDN-F \cite{eccv18refrmv_BDN} & RmNet~\cite{wen2019single} & ERRNet-F~\cite{wei2019single_ERR} & IBCLN~\cite{li2020single} & YTMT-UCS~\cite{hu2021trash}& \textbf{PRRN+RPEN}  \\ \hline 
          PSNR & 19.56 & 18.92 & 19.36 & 22.18 & {23.57} &  \textcolor{blue}{23.85}&\textcolor{red}{25.30}    \\
          SSIM & 0.736 & 0.737 & 0.725 & 0.756 & {0.783} & \textcolor{blue}{0.810}& \textcolor{red}{0.884}    \\ 
          \bottomrule[1pt]
      \end{tabular}
  }
    \caption{Experimental comparison on the Nature dataset using the IBCLN strategy. The best results are highlighted in \textcolor{red}{red}, and the second best results are indicated in \textcolor{blue}{blue}.}
                        \label{table:ibcln}
\end{table}

\subsection{Ablation study}
\begin{figure*}[t!]
\centering
\vspace*{-7mm}
\begin{subfigure}{0.5\linewidth}
  \centering
  \includegraphics[scale=0.5]{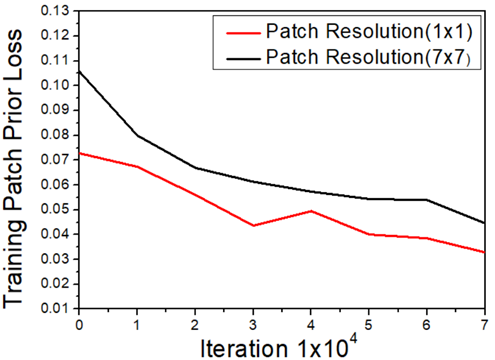}
  \subcaption{Variation of training loss.}
\end{subfigure}%
\begin{subfigure}{0.5\linewidth}
  \centering
  \includegraphics[scale=0.5]{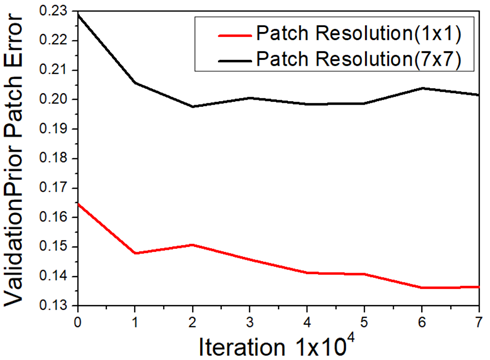}
  \subcaption{Variation of validation patch pixel error.}
\end{subfigure}
\caption{
Training loss and validation patch pixel error with different segmentation patch counts. A denser segmentation approach leads to training difficulties and a decrease in accuracy on the validation set.}
\label{fig:patcherror}
\end{figure*}

\begin{figure*}[t!]
\centering
\vspace*{-4mm}
\begin{subfigure}{0.16\linewidth}
  \centering
  \includegraphics[scale=0.68]{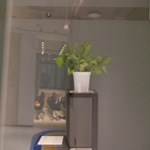}
  \subcaption{Input}
\end{subfigure}
\begin{subfigure}{0.16\linewidth}
  \centering
  \includegraphics[scale=0.68]{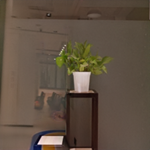}
  \subcaption{1$\times$1 }
\end{subfigure}
\begin{subfigure}{0.16\linewidth}
  \centering
  \includegraphics[scale=0.68]{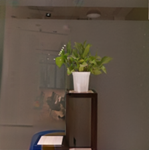}
  \subcaption{7$\times$7}
\end{subfigure}
\begin{subfigure}{0.16\linewidth}
  \centering
  \includegraphics[scale=0.68]{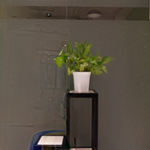}
  \subcaption{14$\times$14}
\end{subfigure}
\begin{subfigure}{0.16\linewidth}
  \centering
  \includegraphics[scale=0.68]{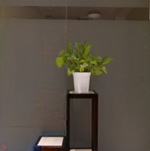}
  \subcaption{28$\times$28}
\end{subfigure}%
\begin{subfigure}{0.16\linewidth}
  \centering
  \includegraphics[scale=0.68]{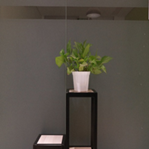}
  \subcaption{GT}
\end{subfigure}%
\caption{Visual comparison of patch resolution demonstrates an improvement in reflection removal effectiveness.}
\label{fig:patchsize}
\end{figure*}

\vspace{-2mm}
\textbf{Ablation Study 1: Effect of Patch Resolution on Reflectance Prior Estimation:} 
As the occurrence of reflections in superimposed images is neither limited to a specific region nor uniform. It is difficult to accurately determine the resolution of reflection intensity prior. 
We train RPEN using a two-segment patch resolution, where the input image is divided into 1x1 and 7x7 patches. We assess the variations in training loss and the patch pixel error on the validation set. As shown in Figure \ref{fig:patcherror}, with 7x7 patches of reflection intensity prior, training becomes significantly more challenging compared to 1x1 patches. 
It exhibits larger losses, difficulty in convergence, and larger validation patch pixel errors. This observation further confirms that increasing the receptive field is beneficial for extracting reflection intensity prior information. On the contrary, as summarized in Table \ref{table:varypatchssim} and Figure \ref{fig:patchsize}, when accurate prior truth is provided as input, the use of smaller patches significantly improves accuracy and the effectiveness of reflection removal.
\begin{table}
  \centering
  \begin{tabular}{ccccccc}
    \toprule
     &Proposed&1$\times$1&7$\times$7 &14$\times$14&28$\times$28 \\
    \midrule
    PSNR &24.31 &26.79  & 27.01 &27.98 &28.97 \\
    SSIM &0.815 &0.831  &0.860  &0.879 & 0.901 \\
    \bottomrule
  \end{tabular}
  \caption{Effect of varying patch resolution in truth reflection intensity prior on reflection removal performance.}
  \label{table:varypatchssim}
\end{table}

\begin{figure*}[t!]
\centering
\vspace*{-4mm}
\begin{subfigure}[t]{0.19\linewidth}
  \centering
  \includegraphics[scale=0.55]{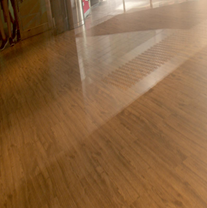}
  \subcaption{Input}
  \label{fig:cam-a}
\end{subfigure}\hfill
\begin{subfigure}[t]{0.19\linewidth}
  \centering
  \includegraphics[scale=0.55]{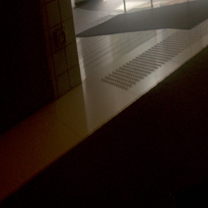}
  \subcaption{Reflection Image}
    \label{fig:cam-b}
\end{subfigure}\hfill
\begin{subfigure}[t]{0.19\linewidth}
  \centering
  \includegraphics[scale=0.55]{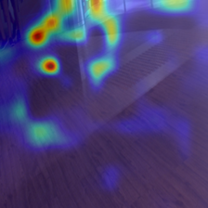}
  \subcaption{\centering PRRN\newline Activation Map}
    \label{fig:cam-c}
\end{subfigure}\hfill
\begin{subfigure}[t]{0.19\linewidth}
  \centering
  \includegraphics[scale=0.55]{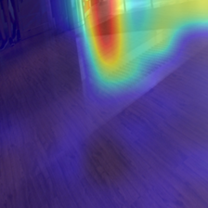}
  \subcaption{\centering RPEN\newline Activation Map}
      \label{fig:cam-d}
\end{subfigure}\hfill
\begin{subfigure}[t]{0.19\linewidth}
  \centering
  \includegraphics[scale=0.55]{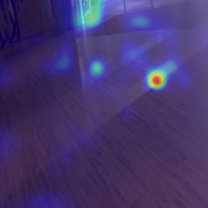}
  \subcaption{\centering PRRN+RPEN\newline Activation Map}
      \label{fig:cam-e}
\end{subfigure}%
\vspace*{-2mm}
\caption{Visual comparison of patch resolution demonstrates the impact of RPEN on the activation maps of the decoder.}
\label{fig:cam}
\end{figure*}

\textbf{Ablation Study 2: Effect of RPEN:} 
As summarized in Table \ref{table:cdr}, we conduct tests on the RPEN by first only using the PRRN on the CDR benchmark, and then incorporating our RPEN module for training and testing. We find that our RPEN significantly improve the performance of the network in the reflection removal, resulting in a 0.87 increase in PSNR. 
To present more intuitive understanding on the effect of RPEN, we utilize LayerCAM (Layer Class Activation Map) \cite{jiang2021layercam} to observe the changes in the gradient activation map in the 4th layer of our U-net decoder. We select an image with reflections occurring in a small region to visualize the changes in attention better as shown in Figure \ref{fig:cam}(\subref{fig:cam-a}).
When the reflection image is presented as shown in Figure \ref{fig:cam}(\subref{fig:cam-b}), we observe that the attention of PRRN is primarily concentrated on non-reflection areas, as depicted in Figure \ref{fig:cam}(\subref{fig:cam-c}). However, with the guidance of the reflection intensity prior provided by RPEN (Figure \ref{fig:cam}(\subref{fig:cam-d})), the model successfully directs its attention to the accurate reflection areas, as depicted in Figure \ref{fig:cam}(\subref{fig:cam-e}). 

\section{Conclusion}
\vspace{-2mm}
We have introduced RPEN and PRRN for reflection intensity prior based SIRR. While our approach achieves state-of-the-art accuracy in real-world benchmarks, it is important to acknowledge that the limited accuracy of reflection intensity prior prediction in some case can be improved with multi-resolutional RPEN in the future.

\bibliographystyle{plain}
\bibliography{han}

\medskip

\end{document}